%% file: main.tex
\documentclass[10pt,twocolumn,letterpaper]{article}

\usepackage[pagenumbers]{cvpr} %

\input{preamble}
\definecolor{cvprblue}{rgb}{0.21,0.49,0.74}
\usepackage[pagebackref,breaklinks,colorlinks,allcolors=cvprblue]{hyperref}
\usepackage[table]{xcolor}

\def\paperID{5189} %
\def\confName{CVPR}
\def\confYear{2026}

\newcommand{\model}{\textbf{Vega}\xspace}

\title{Vega: Learning to Drive with Natural Language Instructions}

\author{
  Sicheng Zuo$^{1, *}$ \quad
  Yuxuan Li$^{1, *}$ \quad
  Wenzhao Zheng$^{1, *, \dagger}$ \quad 
  Zheng Zhu$^{2}$ \quad
  Jie Zhou$^{1}$ \quad
  Jiwen Lu$^{1}$
  \vspace{2mm} \\
  $^1$Tsinghua University \quad $^2$GigaAI \\
  Project Page: \url{https://zuosc19.github.io/Vega}\\
  Large Driving Models: \url{https://github.com/wzzheng/LDM}
}

\begin{document}

\twocolumn[{%
\renewcommand\twocolumn[1][]{#1}%
\vspace{-12mm}
\maketitle
\vspace{-10mm}
\begin{center}
    \centering
    \includegraphics[width=\linewidth]{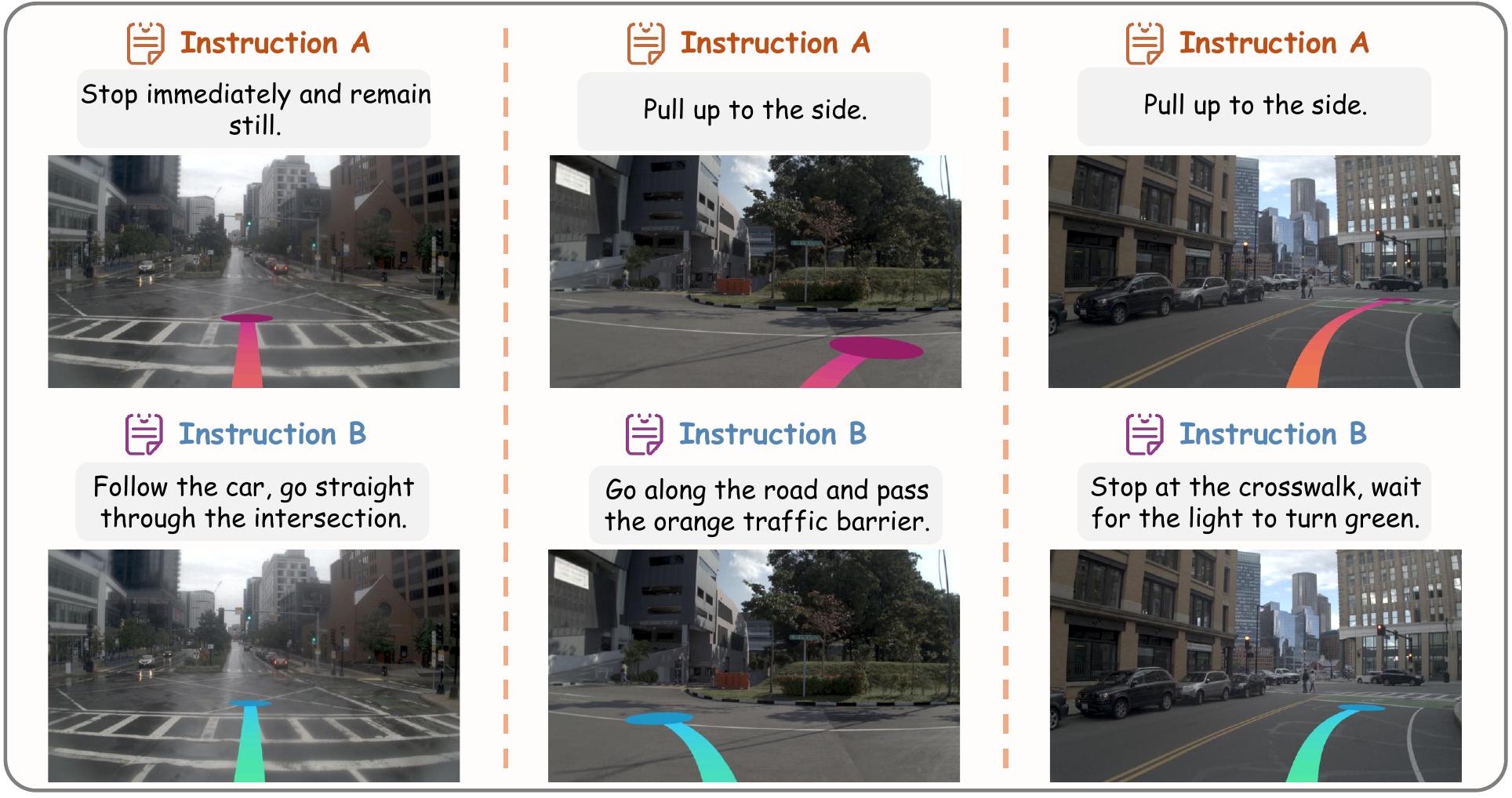}
    \vspace{-7mm}
    \captionof{figure}{\textbf{Visualizations of our model for instructional driving.}
We propose a unified vision-language-world-action model, Vega, for instruction-based generation and planning. Vega can predict multiple trajectories in the same scenario following diverse instructions.
}
\label{fig:teaser}
\end{center}%
\vspace{-2mm}
}]

\begingroup
\renewcommand\thefootnote{}
\footnotetext{
  $^*$Equal contributions.
  $^\dagger$Project leader.
}
\endgroup

\input{arxiv/sec/0_abstract}
\input{arxiv/sec/1_introduction}
\input{arxiv/sec/2_related_work}
\input{arxiv/sec/3_method}
\input{arxiv/sec/4_experiments}
\input{arxiv/sec/5_conclusion}

{
    \small
    \bibliographystyle{ieeenat_fullname}
    \bibliography{main}
}

\end{document}

%% file: arxiv/sec/0_abstract.tex
\begin{abstract}
Vision-language-action models have reshaped autonomous driving by incorporating natural language into the decision-making process.
However, most existing pipelines only utilize the language modality for scene descriptions or reasoning and lack the flexibility to follow diverse user instructions for personalized driving. 
To address this, we first construct a large-scale driving dataset (\textbf{InstructScene}) containing around 100,000 scenes annotated with diverse driving instructions with the corresponding trajectories. 
We then propose a unified \textbf{vision-language-world-action} model, \model, for instruction-based generation and planning.
We employ the autoregressive paradigm to process visual inputs (\textbf{vision}) and language instructions (\textbf{language}) and the diffusion paradigm to generate future predictions (\textbf{world} modeling) and trajectories (\textbf{action}). 
We perform joint attention to enable interactions between the modalities and use individual projection layers for different modalities for more capabilities. 
Extensive experiments demonstrate that our method not only achieves superior planning performance but also exhibits strong instruction-following abilities, paving the way for more intelligent and personalized driving systems.
Code is available at \href{https://github.com/zuosc19/Vega}{{\texttt{https://github.com/zuosc19/Vega}}}.
\vspace{-1mm}
\end{abstract}

%% file: arxiv/sec/1_introduction.tex
\section{Introduction}
\label{sec: intro}
Vision-centric autonomous driving is a promising direction due to its economic advantages and scalability~\cite{bevformer, tpvformer, surroundocc, occ3d, uniad, vad}. 
Conventional methods typically follow a modular pipeline of perception~\cite{bevdet, bevfusion, gaussianformer, gaussianformer2, quadricformer}, prediction~\cite{occworld, occllama, gaussianworld, doe-1}, and planning ~\cite{uniad, vad, genad, vadv2, gaussianad}, which heavily relies on expensive 3D annotations and thus faces limitations in real-world applications. 
Recently, vision-language-action (VLA) models have emerged to leverage rich world knowledge from large language models to map visual inputs to driving actions~\cite{omnidrive, emma, openemma, orion}, demonstrating remarkable generalization across driving scenarios.

Despite their good generalization across driving scenarios, most existing VLA models only use languages for scene descriptions or decision reasoning and lack flexible instruction-following capabilities~\cite{opendrivevla, orion, autovla, drivemoe, recogdrive}.
They are either trained to imitate an averaged expert policy, or are confined to a closed set of simple navigational commands like ``turn left'' or ``go straight'', failing to generalize to open-ended and flexible natural language instructions.
In contrast, a general driving agent should not only navigate autonomously but also comprehend and execute diverse, user-specified natural language instructions. 
For instance, a user in a hurry might instruct the vehicle to ``overtake the front car to catch the next green light'' rather than adhere to the conservative policy learned from the training data. 

To facilitate the shift from imitation driving to instructional driving, we construct a large-scale driving dataset, \textbf{InstructScene}, with around 100,000 instruction-annotated scenes and the corresponding trajectories built on NAVSIM~\cite{navsim-v1}. 
While a direct way is to train a VLA model on our driving dataset containing rich instructions, we find that it struggles to generate feasible trajectories and follow instructions accurately.
We think this is due to the significant information disparity between the high-dimensional visual-instruction inputs and the low-dimensional action prediction, making it difficult for the model to learn a generalizable mapping from high-level instructions to low-level actions in complex and dynamic environments. 

To address this, we propose a unified \textbf{vision-language-world-action} model, \model, for joint instruction-based generation and planning.
We train the model to jointly perform future image generation and action planning  conditioned on past observations and language instructions. 
This task provides a dense and pixel-level supervision signal, compelling the model to learn the causal relationships among instructions, actions, and visual predictions. 
The joint modeling enforces consistency between predictions, enabling mutual supervision and refinement. 
Our model adopts a mixed autoregressive-diffusion transformer architecture~\cite{MoT, janusflow, llamafusion, bagel} to achieve unified \textbf{vision}-\textbf{language} understanding, \textbf{world} modeling, and \textbf{action} planning. 
Specifically, we use the autoregressive pipeline for visual and instruction understanding, and the diffusion pipeline~\cite{rectified-flow, flow-matching} for image and action generation.
We use joint attention to enable interactions across all modalities and employ a Mixture-of-Transformers (MoT) design~\cite{MoT} to effectively decouple the parameters associated with different modalities and enhance the model capacity for joint generation and planning. 
Extensive experiments on the NAVSIM~\cite{navsim-v1, navsim-v2} benchmark show that our model not only achieves superior planning performance but also demonstrates a remarkable ability to generate high-fidelity and instruction-compliant future images and plausible trajectories.

%% file: arxiv/sec/2_related_work.tex
\section{Related Work}
\label{sec: related_work}

\subsection{VLM and VLA for Autonomous Driving}
\label{sec: related_work_vlm}
The extensive world knowledge and reasoning capabilities of vision-language models (VLMs) have driven their applications in autonomous driving~\cite{lmdrive, drivemm, drivegpt4, gpt-driver}. 
Early works primarily leveraged VLMs for high-level driving scene understanding and reasoning, but could not output drivable trajectories~\cite{dolphins, drivemm, drivelm, nuscenes-qa, nuinstruct, nuplanqa, lingoqa, alphadrive}. 
Subsequent methods attempted to have VLMs directly predict textual waypoints~\cite{omnidrive, hint-ad, emma, openemma}, but they struggled due to the inherent limitations of LLMs in precise numerical reasoning~\cite{math-chatgpt, mathbert}. 
This led to the development of VLA models, which integrate a planning module for end-to-end trajectory prediction~\cite{senna, opendrivevla, orion}. 
Common planning approaches include autoregressive prediction of discretized waypoints~\cite{opendrivevla, diffvla, autovla}, diffusion-based trajectory generation~\cite{drivemoe, orion, recogdrive}, and direct regression via an MLP head~\cite{simlingo}. 
However, these models suffer from sparse action supervision and often rely on auxiliary understanding and reasoning tasks to guide the learning process~\cite{opendrivevla, orion, autovla}. 
In contrast, \model employs world modeling to provide a dense signal to enhance instruction-based planning.

\subsection{World Models for Autonomous Driving}
\label{sec: related_work_world_model}
World models are typically defined as generative models that predict future states conditioned on past observations and current actions~\cite{world-models}. 
In autonomous driving, applications of world models can be categorized into three main approaches: image-based, occupancy-based, and VLA-based methods. Image-based methods leverage powerful generative architectures to synthesize high-fidelity driving videos, primarily for data generation and scene simulation~\cite{gaia-1, gaia-2, drivedreamer-1, drivedreamer-2, vista, drive-wm}. 
Occupancy-based methods model scene evolution in 3D occupancy space to enhance scene understanding~\cite{occllama, driveworld, gaussianworld} and planning~\cite{occworld, occllama, drive-occworld, preworld}, but their reliance on dense 3D labels limits scalability. 
Recently, VLA-based methods have emerged with Doe-1~\cite{doe-1} first proposing a closed-loop driving model that unifies scene understanding, prediction, and planning. 
DriveVLA-W0~\cite{drivevla-w0} integrated world modeling into a VLA framework to provide dense supervision and enhance planning. 
However, they can not perform instruction-based prediction and planning. 
Our work enables this capability, allowing the model to predict corresponding future scenes and driving trajectories conditioned on flexible language instructions.

\subsection{Unified Visual Understanding and Generation}
\label{sec: related_work_unifed}
Unified visual understanding and generation methods can be categorized into three main pipelines: quantized autoregressive (AR), external diffusion, and integrated transformers. 
Quantized AR models quantize images into discrete tokens~\cite{argen, vqgen}, enabling generation within the native autoregressive framework~\cite{janus, janus-pro, unified-io-2, tokenflow, liquid, vila-u, chameleon, emu3}. 
While this design is straightforward, its visual quality typically lags behind that of diffusion-based methods. 
The External Diffuser approach pairs a VLM with an external diffusion model~\cite{dreamllm, seed-x, emu2, metamorph}. 
The VLM provides a high-level understanding by generating a few latent tokens that condition the diffusion generator. 
However, this narrow interface between understanding and generation can restrict information flow~\cite{bagel}. 
Integrated transformer models merge autoregressive and diffusion mechanisms into a single transformer \cite{janusflow, llamafusion, transfusion_diffuse, bagel, MoT}, enabling a deep integration of powerful understanding and generation capabilities. 
In this paper, we adopt the integrated transformer to achieve instruction-based joint visual generation and action planning.

%% file: arxiv/sec/3_method.tex
\section{Proposed Approach}
\label{sec: method}

\begin{figure*}[t]
\begin{center}
    \centering
    \includegraphics[width=\linewidth]{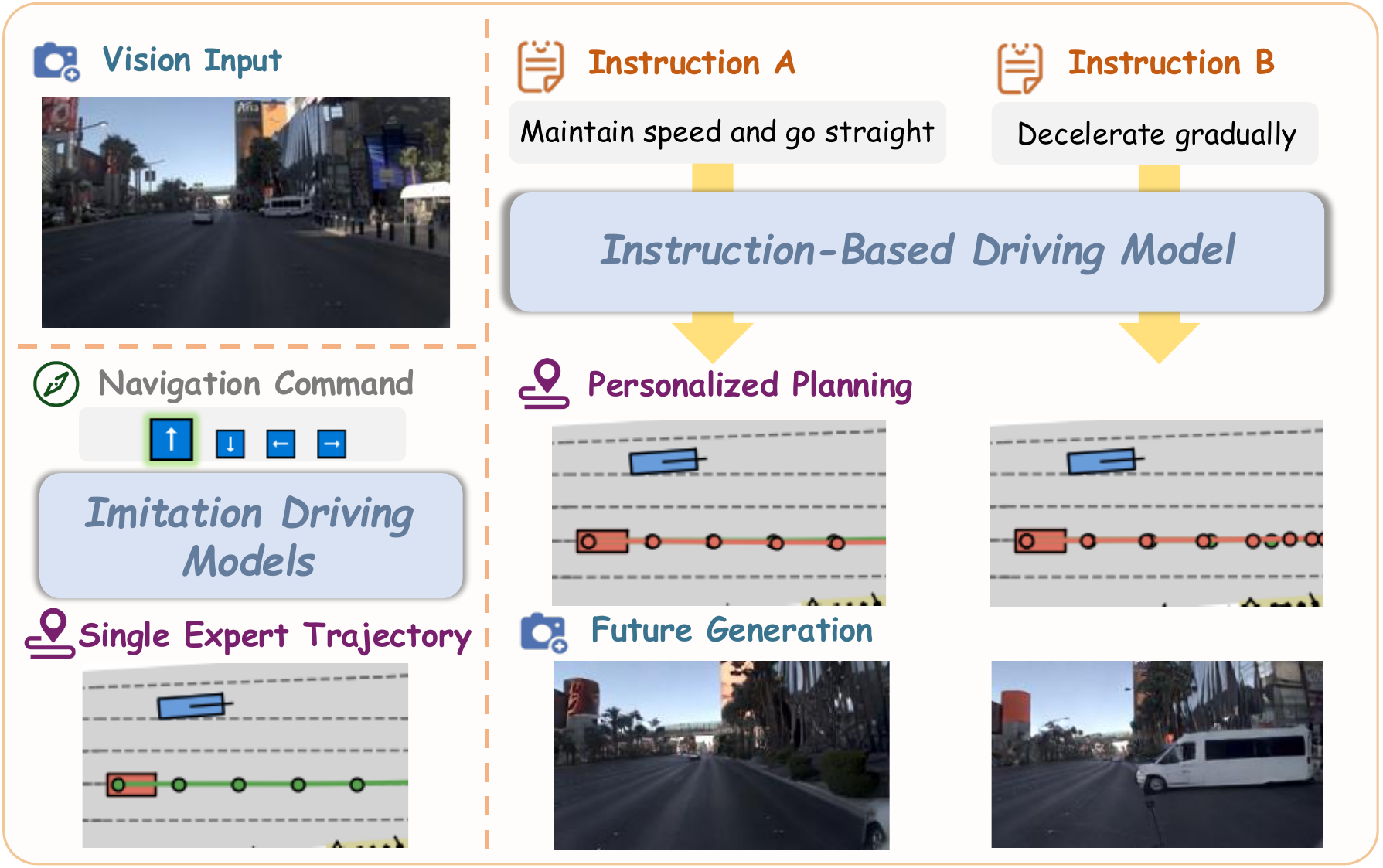}
    \vspace{-7mm}
    \captionof{figure}{\textbf{Overview of our model.}
Compared to traditional imitation driving models, which can only predict the single expert trajectory, \model can follow natural language instructions to generate diverse planning trajectories and future image predictions. 
}
\label{fig:compare}
\end{center}%
\vspace{-8mm}
\end{figure*}

\subsection{Imitation Driving to Instructional Driving}
An autonomous driving model $\mathcal{M}$ usually takes as input the past $T$ and current image observations $[I_{t-T}, \dots, I_{t}]$ and past $T$ actions $[A_{t-T}, \dots, A_{t-1}]$, and predicts the current action $A_t$ for the ego car, which can be formulated as:
\begin{equation}
    A_t = \mathcal{M}([I_{t-T}, \dots, I_{t}], [A_{t-T}, \dots, A_{t-1}]).
\end{equation}

Conventional methods often adopt a perception-prediction-planning pipeline.
The perception module $\mathcal{P}_{er}$ extracts the scene representation $\mathbf{z}$ from observations $[I_{t-T}, \dots, I_{t}]$. Then the prediction module $\mathcal{P}_{re}$ forecasts the future motion $\mathbf{v}$ of agents based on $\mathbf{z}$. Finally, the planning module $\mathcal{P}_{lan}$ uses $\mathbf{z}$, $\mathbf{v}$, and historical ego actions $[A_{t-T}, \dots, A_{t-1}]$ to plan the current ego action $A_t$. This multi-step pipeline can be expressed as:
\begin{align}
    \mathbf{z} &= \mathcal{P}_{er}(I_{t-T}, \dots, I_{t}]), \\
    \mathbf{v} &= \mathcal{P}_{re}(\mathbf{z}), \\
    A_t &= \mathcal{P}_{lan}(\mathbf{z}, \mathbf{v}, [A_{t-T}, \dots, A_{t-1}]). 
\end{align}
However, such methods heavily rely on costly high-quality 3D annotations, which greatly limits their scalability.

Recently, vision-language-action (VLA) models have been applied to autonomous driving, leveraging their rich world knowledge and demonstrating strong generalization across diverse scenarios.
Based on past observations $[I_{t-T}, \dots, I_{t}]$ and historical actions $[A_{t-T}, \dots, A_{t-1}]$, current VLA models $\mathcal{W}$ often predict both the textual description of the scene $D$ and the current ego action $A_t$. This end-to-end planning process can be formulated as:
\begin{equation}
    A_t, D_t = \mathcal{W}([I_{t-T}, \dots, I_{t}], [A_{t-T}, \dots, A_{t-1}]).
\end{equation}

Although existing VLA models show remarkable generalization, they fall short in flexible instruction-following. 
Most VLA models are trained to imitate an averaged expert policy or process a closed set of simple navigational commands, failing to handle open-ended natural language instructions.
To address this, we introduce an instruction-based driving model $\mathcal{V}$, which predicts the current ego action $A_t$ based on observations $[I_{t-T}, \dots, I_{t}]$, historical actions $[A_{t-T}, \dots, A_{t-1}]$ and the current user instruction $L_t$. This process can be expressed as:
\begin{equation}
    A_t = \mathcal{V}([I_{t-T}, \dots, I_{t}], [A_{t-T}, \dots, A_{t-1}], L_t).
\end{equation}

To enable instruction-based driving, we constructed a large-scale driving dataset with around 100,000 instruction-annotated scenes based on NAVSIM~\cite{navsim-v1}, where we generated instructions automatically using VLM, supplemented by rule-based methods. 
For each timestep t, we prompt a powerful VLM~\cite{qwen2_5} with future observations $[I_{t+1},\dots, I_{t+N}]$ and actions $[A_{t+1},\dots, A_{t+N}]$ to produce a high-level instruction $L_t$ describing the driving intent of the current ego-vehicle.
This process yields a sequence of image, instruction, and action triplets: $\mathcal{D} = \{\langle I_t, L_t, a_t \rangle\}_{t=1}^{T_{max}}$.
We then train our model on this dataset, equipping it with instruction-following driving capabilities.

\subsection{Unified Generation and Planning}
\label{sec: method_task}
While a direct way to achieve instruction-based driving is to train a VLA model on our driving dataset containing rich instructions, we find that it struggles to generate feasible trajectories and accurately follow instructions, due to the sparse action supervision.
To address the supervision gap, we introduce the vision-language-world-action model, a novel framework that jointly learns instruction-based action planning and future image generation. 
Our core insight is that future image generation provides a dense, pixel-level supervision signal, which helps the model learn the underlying dynamics of the world. 
By joint modeling generation and planning, the model is compelled to learn the causal relationships among instructions, actions, and visual outcomes, which is critical for instruction-based planning.

The framework is formulated as a generative model trained on triplets of images, instructions, and actions, which models the fundamental causal chain of driving: 
An agent perceives the world $I_t$, receives the instruction $L_t$, decides on an action $A_t$, and observes the next outcome $I_{t+1}$. 
At each timestep $t$, the model receives the current observation $I_t$ and instruction $L_t$, and the historical observations $[I_{t-T}, \dots, I_{t-1}]$. It then jointly predicts the action $A_t$ to be executed and the resulting next step $I_{t+1}$.
We apply causal attention modeling to the model's architecture, ensuring that it learns the correct reasoning pathway from instruction to action and then to visual outcome, providing a solid foundation for resolving the supervision gap.

\subsection{Joint Autoregressive-Diffusion Architecture}
\label{sec: method_model}
Unified generation and planning requires our model to not only possess significant visual-text understanding, visual generation, and action planning capabilities, but also integrates them to solve complex driving scenarios. 
Current research mainly follows three approaches to bridge the gap between visual-text-understanding, which primarily uses auto-regressive VLM, and image generation, which often adopts diffusion models. 
However, most methods fall short of our requirements. 
Autoregressive visual generation models with discrete visual tokenizers struggle to match diffusion models in image quality and also suffer from high latency due to their sequential generation pipeline. 
LLMs combined with external diffusers yield competitive results, but are constrained by an information bottleneck caused by the limited number of latent tokens passed from LLMs to generation modules. 
To address these, we adopt the Integrated Transformer architecture~\cite{bagel}, which fuses auto-regressive VLM and diffusion transformer into a single model, enabling the generation module to interact with the understanding module without information loss and resulting in unified understanding and generation capabilities. 

Our integrated model employs a unified paradigm to predict images and actions. It first encodes multi-modal inputs, including text, images, and actions, and concatenates them to the noises of target images or actions, forming a unified sequence. 
The model then processes the sequence as a whole, calculating causal attention across modalities to ensure full information flow among text, image, and action latents. 
Finally, the denoised latents are decoded by their respective decoders into images or actions. 
 
\textbf{Encoding Inputs}. 
To prepare the multi-modal inputs for the forward pass, we first encode them with corresponding tokenizers. For text, we tokenize natural language inputs $L_t$ with the Qwen2.5 tokenizer. For visual understanding, we only use the forward-view camera images as visual observations, which are encoded by a VAE encoder into latents $F_t^V$. To enrich the visual context, we also encode input images with a SigLIP2 ViT encoder, and append the latents to the corresponding image's VAE latents. For action, we first convert the 2D absolute trajectory $traj=[(x, y, \theta), \dots]$ into relative movements between consecutive steps $A=(\Delta x, \Delta y, \Delta \theta)$, so that actions from different steps share a distribution and can be easily normalized. We project the normalized relative action sequence into the latent dimension of the model with a linear head.

\begin{figure}[t]
    \centering
    \includegraphics[width=0.475\textwidth]{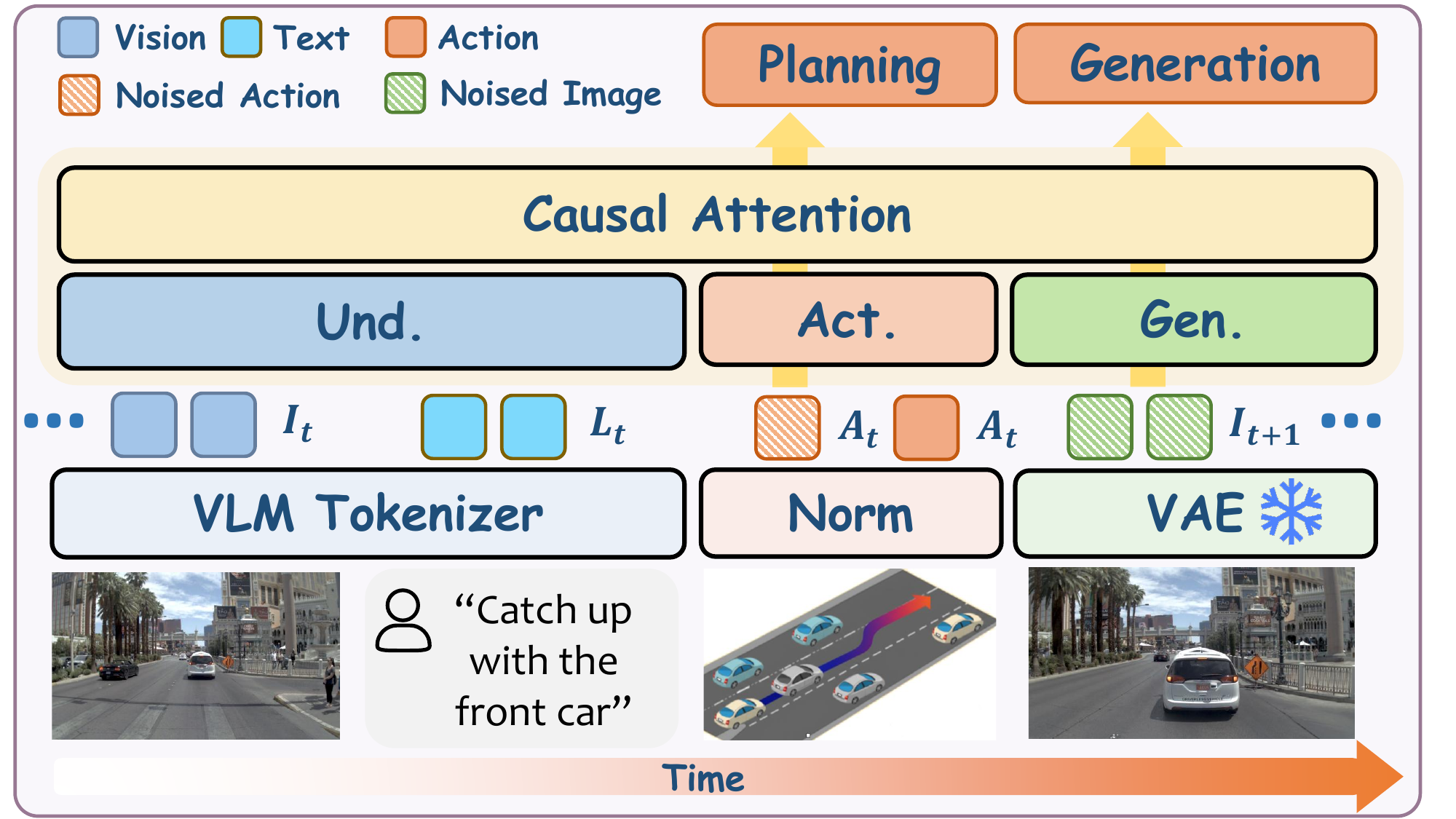}
    \vspace{-7mm}
    \caption{\textbf{Framework of our Unified Vision-Language-World-Action Model.}
    We jointly model action planning and image generation using multi-modal inputs and a MoT architecture \cite{MoT}. 
    }
\label{fig:framework}
\vspace{-5mm}
\end{figure}

\textbf{Constructing Input Sequence}. 
We then combine the multi-modal segments in an interleaving manner. 
The historical images $[I_{t-T}, ..., I_t]$ and actions $[A_{t-T}, ..., A_{t-1}]$ are placed at the beginning, followed by natural language instructions $L_t$. 
When performing the action planning task, we then append a noisy target action $A_t^{noisy}$. 
Otherwise, we first add the ground truth current action $A_t$, then append a noisy future image $I_{t+K}^{noisy}$ for visual generation. 
Due to the strictly causal nature of our sequence $S=[I_0, L_0, A_0, ..., I_n, L_n, A_n]$, we set the attention mask as a blocked lower triangular matrix, so that each block, representing an image, action, or instruction, can only attend to previous blocks. 
In the text block, we adopt a strictly lower triangular mask for causal self-attention and allocates consecutive RoPE indices to textual tokens. 
In the image or action block, we adopt a full attention mask and share the same RoPE index for all tokens, using sinusoidal positional embedding to encode relative position instead. 

During inference, the model denoises the action and future image sequentially, where future image prediction is conditioned on a fully denoised action. 
While during training, the two tasks are optimized jointly for training efficiency. 
A direct concatenation of noisy action and image inputs would cause later tokens to attend to noisy preceding latents, creating a mismatch with inference and degrading training. 
To resolve this, we duplicate each latent that serves both as a prediction target and as a condition for subsequent predictions. 
Specifically, we add noise to the first copy $F_t^{noisy}$ and use it for denoising supervision, while keeping the second copy $F_t^{clean}$ as the condition input. 
We further mask $F_t^{noisy}$ from all subsequent tokens, ensuring that they attend only to the clean latents. 
This design allows us to train multiple diffusion processes within a single autoregressive sequence efficiently.

\textbf{Integrated Transformer}.
To enhance the performance of our integrated transformer, we decouple the modules and weights in charge of each capability so that they can be optimized individually. 
Unlike the Mixture of Expert (MoE) technique, which only uses separate weights for FFN, we employ the Mixture of Transformers (MoT) architecture~\cite{MoT, llamafusion}, where all trainable parameters of the transformer, including attention and FFN layers, are duplicated for each module. 
This design has been shown to not only converge faster, but also maintain higher model capacity~\cite{bagel}. 
Specifically, we process visual and text understanding tokens with a understanding transformer based on Qwen2.5 LLM~\cite{qwen2_5}, which has a hidden size of 3584 and a depth of 28 layers. 
Image generation tokens are processed by a generation transformer of the same design. The weights of both transformers are initialized from Bagel-7B~\cite{bagel}. 
Due to the relatively low dimensionality of the action space, we reduce the hidden size of the action module to 256, thus reducing action-related computation without significantly degrading model performance. 

During the forward process, the interleaving multi-modal sequence is split into segments and passed onto their respective modules in each attention and FFN layers.
To calculate global causal attention, the sequence is re-assembled to be processed as a whole. 
Tokens for image generation and action planning are then extracted from the output sequence for final prediction.

\subsection{Training and Inference}
\label{sec: method_train}
 We implement a single-stage training paradigm to cover both action planning and world modeling. For action planning, we train the model to predict the action plan $A_t^{(N)}=[A_t, ..., A_{t+N-1}]$ based on past observations $I_t^{(-T)}=[I_{t-T}, ..., I_t]$ and current driving instruction $L_t$. For world modeling, we train the model to predict future image observation $I_{t+N}$ based on past images $I_t^{(-T)}$, current driving instruction $L_t$ and action plan $A_t^{(N)}$. We use the MSE of the normalized relative action $A$ as action loss:
 \begin{equation}
    \mathcal{L}_A = \mathbb{E}_{A_t^{(N)}, \epsilon, m}[ || \epsilon - \hat\epsilon(A_t^{(N)}, \epsilon, m, I_t^{(-T)}, L_t) ||^2 ], 
\end{equation}
where $\epsilon\sim\mathcal{N}(0, \mathbf{I})$ is sampled Gaussian noise and $m$ is a random timestep, and MSE of the VAE latents $F^V$ as image loss: 
 \begin{equation}
    \mathcal{L}_V = \mathbb{E}_{F_{t+N}^V, \epsilon, n}[ || \epsilon - \hat\epsilon(F_{t+N}^V, \epsilon, n, I_t^{(-T)}, L_t, A_t^{(N)}) ||^2 ], 
\end{equation}
where $\epsilon\sim\mathcal{N}(0, \mathbf{I})$ is sampled Gaussian noise and $n$ is a random timestep. 
To enable Classifier-Free Guidance (CFG)~\cite{cfg} in inference, we randomly drop text, ViT, clean VAE, and clean action tokens during training. Tokens of the same modality that belong to different images or actions are dropped or kept jointly. 

\input{arxiv/tables/navsim_v2}

In the training stage, we optimize a joint objective with loss $\mathcal{L} = \lambda_A \cdot \mathcal{L}_A + \lambda_V \cdot \mathcal{L}_V$. This allows the model to learn world knowledge alongside planning capabilities. 
In the inference stage, we use Classifier-Free Guidance Diffusion~\cite{cfg} to generate actions, with both image guidance and text guidance enabled. 
While we primarily focus on the action planning task during inference, the model retains its image generation capabilities from the training stage. 

%% file: arxiv/tables/navsim_v2.tex
\begin{table*}[t!] \small
\centering
\caption{\textbf{Comparison with state-of-the-art methods on the NAVSIM v2 with extended metrics.} 
NC: No at-fault Collision. 
DAC: Drivable Area Compliance. 
DDC: Driving Direction Compliance. 
TLC: Traffic Light Compliance. 
EP: Ego Progress. 
TTC: Time to Collision. 
LK: Lane Keeping. 
HC: History Comfort. 
EC: Extended Comfort.
EPDMS: Extended Predictive Driver Model Score.
\dag: Using the best-of-N (N=6) strategy following~\cite{autovla}.
}
\vspace{-3mm}
\label{tab:sota_navsim_v2}
\setlength{\tabcolsep}{8pt}
    \begin{tabular}{l|cccc|ccccc|>{\columncolor{gray!25}}c}
    \toprule
        \textbf{Method} & \textbf{NC $\uparrow$} & \textbf{DAC $\uparrow$} & \textbf{DDC $\uparrow$} & \textbf{TLC $\uparrow$} & \textbf{EP $\uparrow$} & \textbf{TTC $\uparrow$} & \textbf{LK $\uparrow$} & \textbf{HC $\uparrow$} & \textbf{EC $\uparrow$} & \textbf{EPDMS $\uparrow$} \\
    \midrule
        Ego Status & 93.1 & 77.9 & 92.7 & 99.6 & 86.0 & 91.5 & 89.4 & 98.3 & 85.4 & 64.0 \\
        TransFuser~\citep{transfuser} & 96.9 & 89.9 & 97.8 & 99.7 & 87.1 & 95.4 & 92.7 & 98.3 & 87.2 & 76.7 \\
        HydraMDP++~\citep{hydra-mdp} & 97.2 & 97.5 & 99.4 & 99.6 & 83.1 & 96.5 & 94.4 & 98.2 & 70.9 & 81.4 \\
        DriveSuprim~\citep{drivesuprim} & 97.5 & 96.5 & 99.4 & 99.6 & \textbf{88.4} & 96.6 & 95.5 & 98.3 & 77.0 & 83.1 \\
        ARTEMIS~\citep{artemis} & 98.3 & 95.1 & 98.6 & 99.8 & 81.5 & 97.4 & 96.5 & 98.3 & - & 83.1 \\
        DiffusionDrive~\citep{diffusiondrive} & 98.2 & 95.9 & 99.4 & 99.8 & 87.5 & 97.3 & 96.8 & 98.3 & \textbf{87.7} & 84.5 \\ 
        DriveVLA-W0 & 98.5 & \textbf{99.1} & 98.0 & 99.7 & 86.4 & 98.1 & 93.2 & 97.9 & 58.9 & 86.1 \\
    \midrule
        \rowcolor{gray!20}
        \model & 98.9 & 95.3 & 99.4 & \textbf{99.9} & 87.0 & 98.4 & 96.5 & 98.3 & 76.3 & 86.9 \\
        \rowcolor{gray!20}
        \model\dag & \textbf{99.2} & 96.6 & \textbf{99.5} & \textbf{99.9} & 87.5 & \textbf{98.7} & \textbf{97.4} & \textbf{98.4} & 84.5 & \textbf{89.4} \\
    \bottomrule
    \end{tabular}
\vspace{-5mm}
\end{table*}

%% file: arxiv/sec/4_experiments.tex
\section{Experiments}
\label{sec: experiments}

\subsection{Datasets and Benchmarks}
\label{sec: datasets_benchmarks}
\begin{itemize}
    \item \textbf{NAVSIM v1} \cite{navsim-v1} filters OpenScene to remove near-trivial and erroneous scenes, reducing the train split size to 85k. During evaluation, NAVSIM v1 runs a non-reactive simulation at 10Hz for 4 seconds, then scores the driving agent with metrics including No at-fault Collision (NC), Drivable Area Compliance (DAC), Time To Collision (TTC), Comfort (Comf.), and Ego Progress (EP). These metrics are aggregated into the Predictive Driver Model Score (PDMS). We use the train split for training and the test split for evaluation.
    \item \textbf{NAVSIM v2} \cite{navsim-v2} improves simulation realism by enabling reactive traffic. It evaluates agents with the Extended Predictive Driver Model Score (EPDMS), adding metrics including Driving Direction Compliance (DDC), Traffic Light Compliance (TLC), Lane Keeping (LK), History Comfort (HC) and Extended Comfort (EC). 
\end{itemize}

\input{arxiv/tables/navsim_v1}

\input{arxiv/tables/ablation_future}

\input{arxiv/tables/ablation_expert}

\subsection{Implementation Details}
\label{sec: implementation}
\textbf{Instruction Annotation}.
The driving instructions in our \textbf{InstructScene} dataset were generated by a fully-automated two-stage annotation pipeline. We select Qwen2.5-VL-72B-Instruct \cite{qwen2_5} as our annotation model for its powerful visual understanding capabilities. The inputs of each scene are 14 consecutive frames captured by the front-view camera at 2Hz, with a resolution of $(1920, 1080)$. The first 4 frames are considered past and current observations, and the last 10 frames are future observations that will not be available to the driving agent in the inference stage. 
\begin{itemize}
    \item \textbf{Stage One: Scene Understanding}. In stage one, we prompt the model with two requests, designed to convert the visual inputs and expected actions of the driving agent into natural language descriptions. We first instruct the model to describe the scene in the first 4 frames and to identify the traffic participants as well as the static objects. We then instruct it to describe the vehicle's driving behavior in the next 10 frames and its interaction with previously observed traffic participants. 
    \item \textbf{Stage Two: Instruction Formulation}. In stage two, we combine the visual inputs with the scene descriptions generated in stage one, and prompt the model to create concise driving instructions that would guide the driving agent to predict the actions described in stage one. 
\end{itemize}
Since VLMs struggle to accurately perceive ego-vehicle motion, we generate supplementary rule-based instructions. We classify scenes using speed, acceleration, and turn rate thresholds, converting them into natural language. While these closed-set instructions lack diversity, they provide precise ego-motion cues. We then use them as auxiliary prompts for the VLM, combining both to generate accurate and diverse driving instructions. With this pipeline, we annotated $85109$ scenes from the NAVSIM train split and $12144$ scenes from the NAVSIM test split. 

\textbf{Training.}
The model is trained on the NAVSIM train split for 200k steps using 8 H20 GPUs. We set the number of historical images to 4, and predict 8 future actions as well as the future image at the end of the actions. We set the learning rate to 2e-5 with 2500 warmup steps, and use a per-device batch size of 1. The weights of action and image loss are $\lambda_A = \lambda_V =1.0$. We also maintain an EMA model with a decay rate of 0.9999, which is saved as checkpoints.

\subsection{Main Results}
\label{sec: main_results}
As shown in Tables~\ref{tab:sota_navsim_v2} and~\ref{tab:sota_navsim_v1}, our model demonstrates competitive performance on both NAVSIM benchmarks. On NAVSIM v2, it scores 86.9 EPDMS without any additional performance-enhancing techniques, which is comparable to SOTA. Using the best-of-N strategy as prior works~\cite{autovla, drivevla-w0}, it achieves top performance on NAVSIM v2, surpassing state-of-the-art methods on several metrics, including Driving Direction Compliance, Traffic Light Compliance, Lane Keeping, and History Comfort. These results suggest that \model has learned robust instruction following capabilities and benefited from future image prediction training. On NAVSIM v1, our model achieves 87.9 PDMS, matching multi-modal BEV methods, and improves to 89.8 with the best-of-N strategy. 
We note that \model achieves lower performance compared to state-of-the-art VLA-based methods on NAVSIM v1. This discrepancy is partially attributed to NAVSIM v1's inbalanced metrics, which disproportionately favor risk-averse policies over alternative, equally valid strategies learned by our model. 
Furthermore, competing VLA-based methods either require supplementary inputs such as multi-view images with high resolutions, or integrate CoT reasoning via additional RL training. 
Critically, these performance-enhancing mechanisms operate independently of our model’s core architecture and may be modularly incorporated without modifying the primary design. 

\begin{figure}[t]
\begin{center}
    \centering
    \includegraphics[width=\columnwidth]{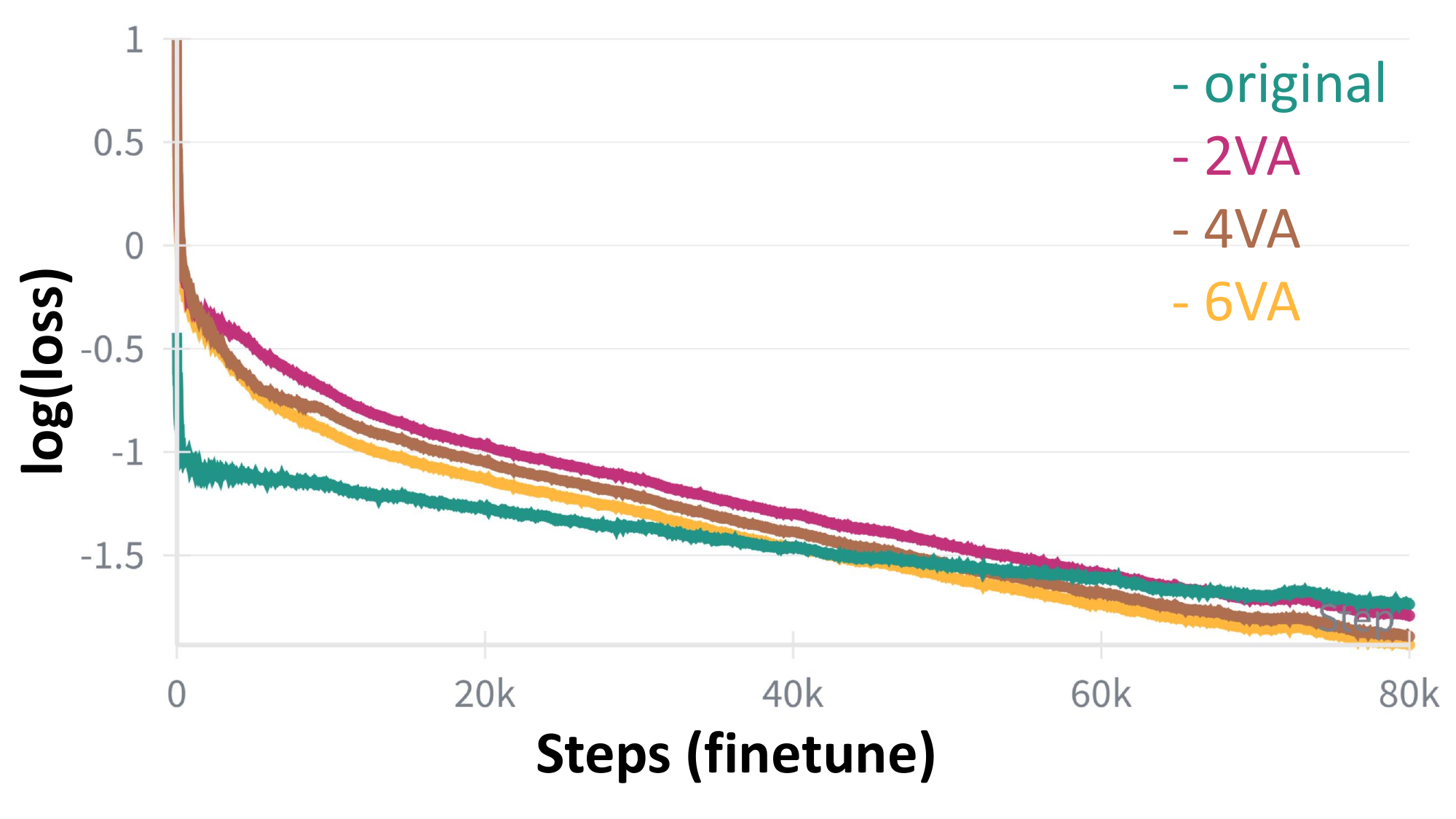}
    \vspace{-8mm}
    \captionof{figure}{\textbf{Ablation of interleaving image-action sequences}.
We compare the training losses of models trained on non-interleaving sequences (original) and interleaving sequences of different lengths.
}
\label{fig:nva}
\end{center}%
\vspace{-10.5mm}
\end{figure}

\begin{figure*}[t]
\centering
\includegraphics[width=1.0\textwidth]{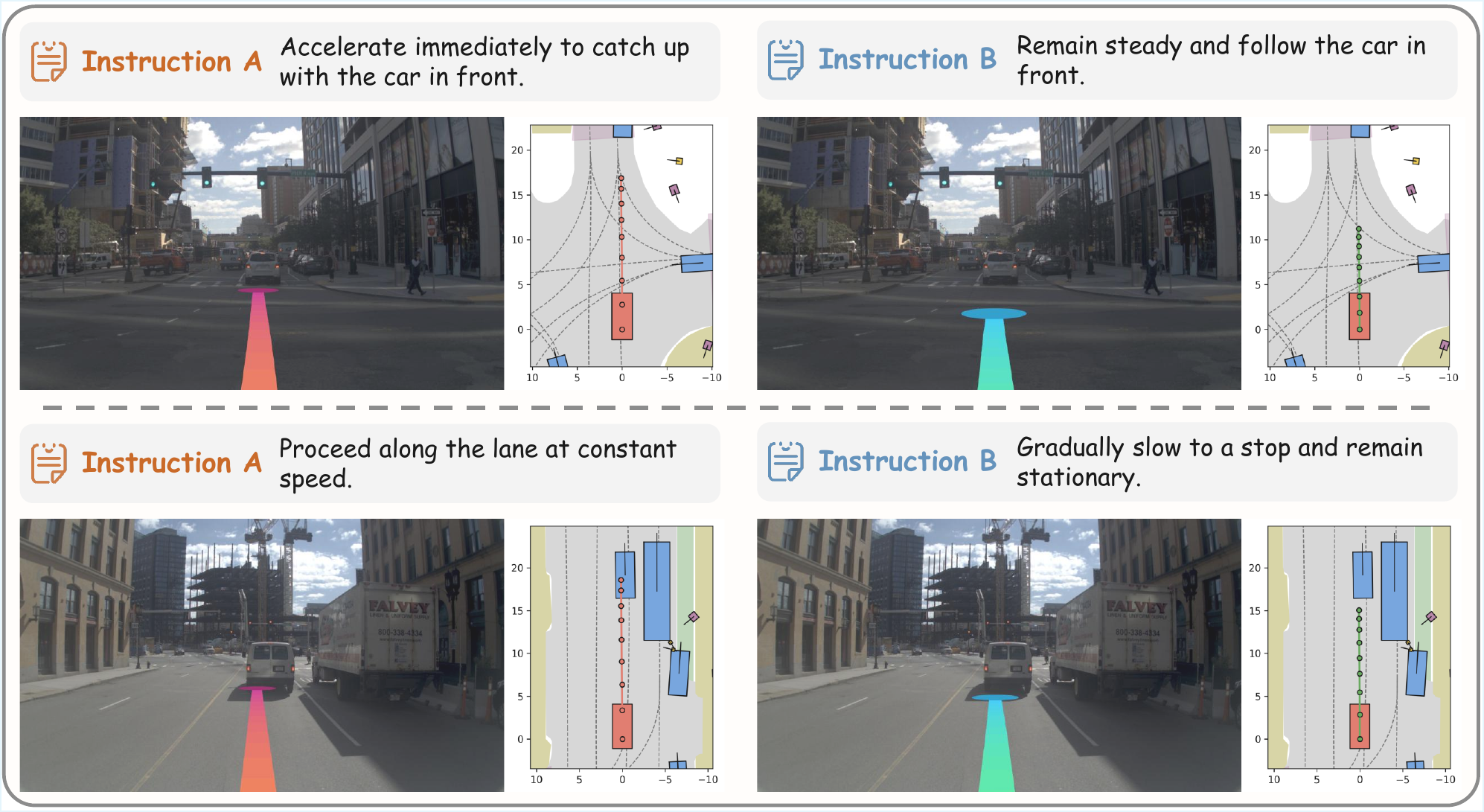}
\vspace{-6mm}
\caption{\textbf{Instruction-based planning examples}.
We visualize the effects of language instructions on action planning with front-view camera images and BEV maps. 
}
\label{fig:additional_examples}
\vspace{-7mm}
\end{figure*}

\subsection{Experimental Analysis}
\label{sec: ablation}
\textbf{Future Frame Prediction}. 
We ablate future frame prediction with three configurations. First, we always predict the next future frame. Second, we randomly sample one of the 8 future frames and specify the chosen index in the text prompt. Third, we remove the future frame prediction task altogether. All three configurations are trained on 8 H20 GPUs, with other hyperparameters identical to training in \Cref{sec: implementation}. The results, shown in \Cref{tab:ablation_frame}, indicate that the task of future frame prediction indeed improves the planning capabilities of the model, but the exact choice of future frame has limited impact on performance. 

\textbf{Interleaving Observation and Action}. 
In our original design, only past images are provided to the model as reference. We argue that interleaving image and action helps the model learn their dynamics, resulting in faster convergence and lower loss during training. Following~\cite{drivevla-w0}, we train the model with interleaving image-action sequences. We ablate the training image-action sequence length with 2, 4 and 6. During training, we interleave the original 4 past images with 3 past actions. Figure~\ref{fig:nva} reveals that although models trained on interleaving sequences suffer from higher loss in the initial stages of training, which can be attributed to the discrepancy between their training designs, they converge significantly faster than models without interleaving sequences, eventually surpassing the latter. In addition, although all interleaving models use the same sequence length during training, those with longer image-action sequences during training show lower losses.

\textbf{Independent Action Module}. 
To validate our design of an additional action expert, we ablate it against using the existing VLM or diffusion modules for action planning. Although these alternatives reduce model size, they increase computational cost because of the high dimensions of these modules. As shown in \Cref{tab:ablation_architecture}, our action expert model slightly outperforms the VLM-module-based planner and significantly surpasses the diffusion-module-based one, confirming the effectiveness of our architecture.

\begin{figure*}[t]
\centering
\includegraphics[page=1,width=0.985\textwidth]{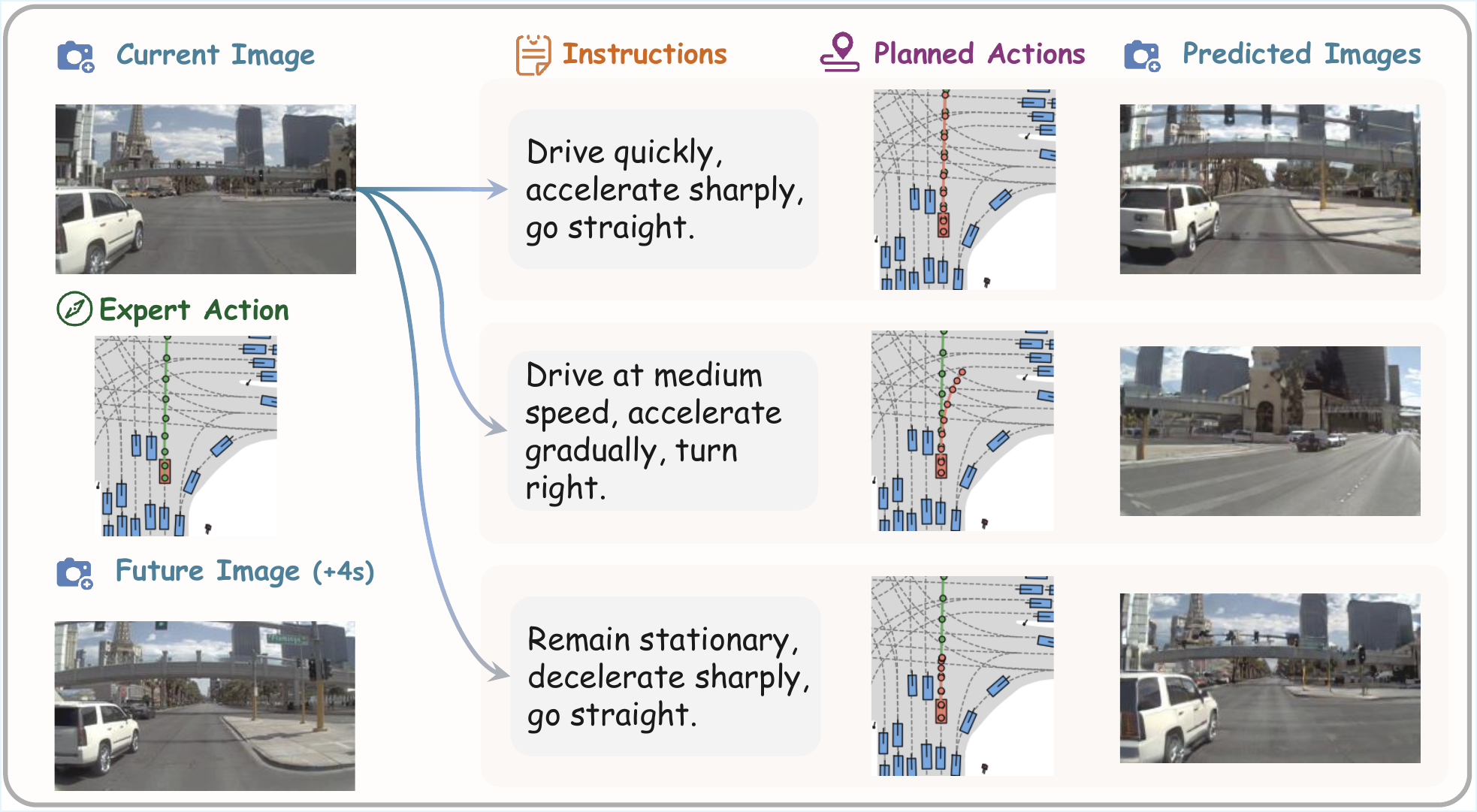}
\includegraphics[page=2,width=0.985\textwidth]{figures/Instruction_cropped.pdf}
\vspace{-2mm}
\caption{\textbf{Future image generation conditioned on instructions and actions.}
In the same scenario, given three sets of instructions, the model plans three action sequences and generates their respective future images. All action sequences follow their instructions and all images are consistent with their actions. 
}
\label{fig:instruction_action}
\vspace{-5mm}
\end{figure*}

\textbf{Visualizations}. 
In addition to Figure~\ref{fig:teaser}, we provide two more examples to visualize the effect of instructions on \model's ability to adjust the vehicle's speed according to user instructions in Figure~\ref{fig:additional_examples}. We test two instructions in each scene and plot the predicted trajectories for the next 8 frames on the front-view-camera image as well as the Bird's-Eye-View (BEV) map. In both scenes, our model successfully increased, decreased, or maintained speed to follow the instructions. 
We also offer a qualitative evaluation of our model's ability to align both its action planning and image generation with user instructions in Figure~\ref{fig:instruction_action}. We select critical scenes where there can be multiple possible courses of action, e.g. approaching the intersection, encountering another vehicle. For each scene, we give the model different sets of instructions, then generate future actions and images sequentially. We observe that \model is able to generate future actions and images that are consistent with the instructions, indicating that our world modeling framework has successfully helped the model learn the dynamics of the driving environment. 

\textbf{VLA Baseline}.
As a straightforward baseline for instructional driving, we extend Qwen-2.5-VL~\cite{qwen2_5} with a planning head to predict future actions based on language instructions. Despite being trained on the same dataset with instruction annotations as \model, this VLA model performs poorly, achieving only $\sim 60$ PDMS and often failing to generate instruction-consistent trajectories. We attribute this limitation to the sparse low-dimensional action supervision, which is insufficient to bridge high-dimensional visual-language inputs and low-level driving actions. This motivates us to explore dense visual supervision from future prediction to improve instruction-based planning.

%% file: arxiv/tables/navsim_v1.tex
\begin{table*}[t] \small
\centering
\caption{\textbf{Comparison with state-of-the-art methods on the NAVSIM v1.} 
NC: No at-fault Collision.
DAC: Drivable Area Compliance.
TTC: Time-To-Collision.
C.: Comfort.
EP: Ego Progress.
PDMS: Predictive Driver Model Score.
Abbreviations: 1x Cam (single front-view camera), Nx Cam (surround-view cameras), L (LiDAR).
 \dag: Using the best-of-N (N=6) strategy following~\cite{autovla}.
 }
 \vspace{-5mm}
\label{tab:sota_navsim_v1}
\setlength{\tabcolsep}{10pt}
\begin{center}
        \begin{tabular}{l|c|c|cc|ccc|>{\columncolor{gray!25}}c}
            \toprule
            \textbf{Method} & \textbf{Ref} & \textbf{Sensors} & \textbf{NC $\uparrow$} & \textbf{DAC $\uparrow$} & \textbf{TTC $\uparrow$} & \textbf{C. $\uparrow$} & \textbf{EP $\uparrow$} & \textbf{PDMS $\uparrow$} \\ 
            \midrule
            Human & - & - & 100 & 100 & 100 & 99.9 & 87.5 & 94.8 \\ 
            \midrule
            \multicolumn{9}{l}{\textit{BEV-based Methods}} \\
            UniAD~\citep{uniad} & CVPR'23 & 6x Cam & 97.8 & 91.9 & 92.9 & \textbf{100.0} & 78.8 & 83.4 \\
            TransFuser~\citep{transfuser} & TPAMI'23 & 3x Cam + L & 97.7 & 92.8 & 92.8 & \textbf{100.0} & 79.2 & 84.0 \\  
            PARA-Drive~\citep{para-drive} & CVPR'24 & 6x Cam& 97.9 & 92.4 & 93.0 & 99.8 & 79.3 & 84.0 \\  
            LAW~\citep{law} & ICLR'25 & 1x Cam & 96.4 & 95.4 & 88.7 & 99.9 & 81.7 & 84.6 \\
            Hydra-MDP~\citep{hydra-mdp} & arXiv'24 & 3x Cam + L & 98.3 & 96.0 & 94.6 & \textbf{100.0} & 78.7 & 86.5 \\
            DiffusionDrive~\citep{diffusiondrive} & CVPR'25 & 3x Cam + L & 98.2 & 96.2 & 94.7 & \textbf{100.0} & 82.2 & 88.1 \\
            WoTE~\citep{wote} & ICCV'25 & 3x Cam + L & 98.5 & 96.8 & 94.4 & 99.9 & 81.9 & 88.3 \\
            \midrule
            \multicolumn{9}{l}{\textit{VLA-based Methods}} \\
            AutoVLA~\citep{autovla} & NeurIPS'25 & 3x Cam & 98.4 & 95.6 & \textbf{98.0} & 99.9 & 81.9 & 89.1 \\
            ReCogDrive~\citep{recogdrive} & arXiv'25 & 3x Cam & 98.2 & \textbf{97.8} & 95.2 & 99.8 & 83.5 & 89.6 \\
            AutoVLA\dag~\citep{autovla} & NeurIPS'25 & 3x Cam & 99.1 & 97.1 & 97.1 & \textbf{100.0} & 87.6 & 92.1 \\
            DriveVLA-W0\dag & arXiv'25 & 1x Cam & \textbf{99.3} & 97.4 & 97.0 & 99.9 & \textbf{88.3} & \textbf{93.0} \\ 
            \midrule
            \rowcolor{gray!20}
            \model & - & 1x Cam & 98.9 & 95.3 & 96.1 & \textbf{100.0} & 81.6 & 87.9 \\ 
            \rowcolor{gray!20}
            \model\dag & - & 1x Cam & 99.2 & 96.6 & 96.9 & \textbf{100.0} & 83.4 & 89.8 \\ 
            \bottomrule
        \end{tabular}
\end{center}
\vspace{-9mm}
\end{table*}

%% file: arxiv/tables/ablation_future.tex
\begin{table}[t] \small
\centering
\setlength{\tabcolsep}{16pt}
\caption{\textbf{Ablation of future image prediction.} 
PDMS: NAVSIM v1 \cite{navsim-v1} benchmark, EPDMS: NAVSIM v2 \cite{navsim-v2} benchmark. }
\vspace{-3mm}
\label{tab:ablation_frame}
        \begin{tabular}{l|c|c}
            \toprule
            \textbf{Setting} & \textbf{PDMS $\uparrow$} & \textbf{EPDMS  $\uparrow$}\\ 
            \midrule
            Random Frame & 77.3 & 75.2  \\ 
            Action Only & 51.8 & 48.9  \\ 
            Next Frame & \textbf{77.9} & \textbf{76.0} \\ 
            \bottomrule
        \end{tabular}
\end{table}

%% file: arxiv/tables/ablation_expert.tex
\begin{table}[!t]\small
\centering
\setlength{\tabcolsep}{16pt}
\vspace{-2mm}
\caption{\textbf{Ablation of action expert.} 
PDMS: NAVSIM v1 \cite{navsim-v1} benchmark, EPDMS: NAVSIM v2 \cite{navsim-v2} benchmark. }
\vspace{-3mm}
\label{tab:ablation_architecture}
        \begin{tabular}{l|c|c}
            \toprule
            \textbf{Setting} & \textbf{PDMS $\uparrow$} & \textbf{EPDMS  $\uparrow$}\\ 
            \midrule
            Use Diffusion & 19.7 & 19.6  \\ 
            Use VLM & 77.6 & 75.7  \\ 
            Action Expert & \textbf{77.9} & \textbf{76.0} \\ 
            \bottomrule
        \end{tabular}
\vspace{-7mm}
\end{table}

%% file: arxiv/sec/5_conclusion.tex
\section{Conclusion}
\label{sec: conclusion}
In this paper, we have aimed to address current driving models' inability to follow diverse driving instructions. 
We have introduced \model, a unified vision-language-world-action model that bridges this gap by leveraging future visual generation as a dense supervision signal. 
We have built a large-scale driving dataset with instruction annotations to enable the training for instructional driving. 
By jointly generating instruction-compliant future images and planning actions, the model learns the causal relationships among instructions, actions, and the visual outcomes. 
Built upon an integrated transformer architecture and an instruction-annotated dataset, our model achieves SOTA planning performance while demonstrating strong instruction-following capabilities in both visual generation and action planning.